\documentclass{article}

\usepackage{microtype}
\usepackage{graphicx}
\usepackage{subfigure}
\usepackage{booktabs} 

\usepackage{hyperref}

\usepackage[accepted]{icml2023-diffxyz}

\usepackage{amsmath}
\usepackage{amssymb}
\usepackage{mathtools}
\usepackage{amsthm}
\usepackage{bm}

\usepackage[capitalize,noabbrev]{cleveref}

\theoremstyle{plain}

\theoremstyle{definition}

\theoremstyle{remark}

\usepackage[textsize=tiny]{todonotes}

\icmltitlerunning{Latent Random Steps as Relaxations of Max-Cut, Min-Cut, and More}

\author{%
  Sudhanshu Chanpuriya \\
  University of Massachusetts Amherst\\
  \texttt{schanpuriya@cs.umass.edu} \\
   \And
  Cameron Musco \\
  University of Massachusetts Amherst\\
  \texttt{cmusco@cs.umass.edu} \\
}

\begin{document}

\twocolumn[
\icmltitle{Latent Random Steps as Relaxations of Max-Cut, Min-Cut, and More}



\icmlsetsymbol{equal}{*}





\vspace{-0.15in}

\makestandardauthors

\vskip 0.35in
]




\begin{abstract}
Algorithms for node clustering typically focus on finding homophilous structure in graphs. 
That is, they find sets of similar nodes with many edges \emph{within}, rather than \emph{across}, the clusters.
However, graphs often also exhibit heterophilous structure, as exemplified by (nearly) bipartite and tripartite graphs, where most edges occur across the clusters. Grappling with such structure is typically left to the task of \emph{graph simplification}.
We present a probabilistic model based on non-negative matrix factorization which unifies clustering and simplification, and provides a framework for modeling arbitrary graph structure.
Our model is based on factorizing the process of taking a random walk on the graph. It permits an unconstrained parametrization, allowing for optimization via simple gradient descent. 
By relaxing the hard clustering to a soft clustering, our algorithm relaxes potentially hard clustering problems to a tractable ones.
We illustrate our algorithm's capabilities on a synthetic graph, as well as simple unsupervised learning tasks involving bipartite and tripartite clustering of orthographic and phonological data.
\end{abstract}

\section{Introduction} \label{sec:intro}

A core method of finding structure in networks is mapping nodes to some smaller set of node clusters based on structural similarity.
There are various algorithms for this task of node clustering, one of the most well-known being the normalized cuts algorithm~\cite{shi2000normalized}, which assigns clusters based on an eigenvector of the normalized graph Laplacian. This algorithm finds a hard clustering, where each node is mapped to exactly one cluster; soft clustering~\cite{yu2005soft} relaxes this problem and instead assigns nodes to clusters probabilistically, so that each node is mapped to a categorical distribution over clusters.

Clustering algorithms generally try to find groups of nodes that are in line with graph homophily, wherein edges connect nodes with similar attributes, and wedges tend to be closed (``a friend of a friend is a friend'').
Only a small number of clustering algorithms can be seen as capturing heterophilous structures, such as (near) bipartiteness: for example, algorithms for the max-cut problem~\cite{grotschel1981weakly} can find approximately bipartite structure in  graphs.
Approaches for the closely related task of graph simplification (also called graph compression) are often more amenable than typical clustering approaches to addressing heterophilous structure. 
Like clustering, simplification is focused on finding structure in graphs, but with the goal of minimizing reconstruction error from a compressed representation. Algorithms for simplification work by merging edges or nodes~\cite{toivonen2011compression, garg2019compression}, or by approximate factorization of the adjacency matrix~\cite{nourbakhsh2014matrix}.

\begin{figure}
    \centering
    \includegraphics[width=0.8\columnwidth]{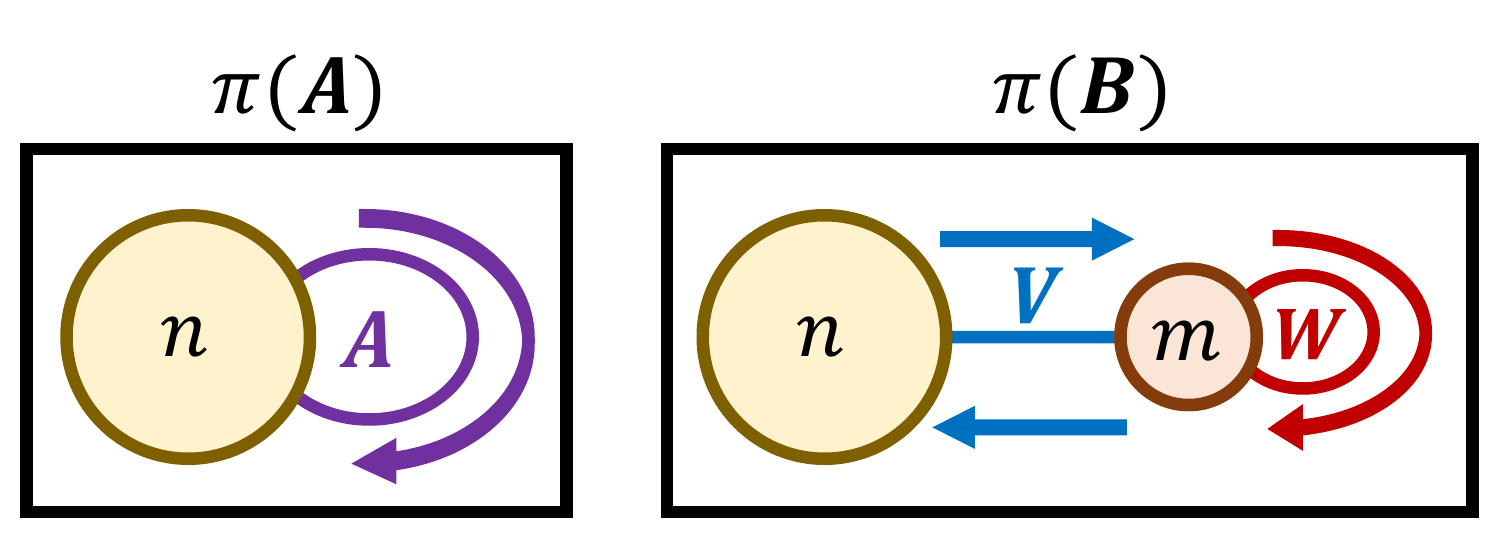}
    \vspace{-0.5em}
    \caption{Diagram of the latent random step model. A random step on a graph $\bm{A}$ with $n$ nodes is approximated by a random step forward on a bipartite graph $\bm{V}$, then a random step on a smaller graph $\bm{W}$ with $m$ nodes, then a random step back on $\bm{V}$.}
    \label{fig:diagram}
    \vspace{-0.9em}
\end{figure}

We present a probabilistic framework which unifies node clustering and graph simplification and is applicable to both homophilous and heterophilous structure. In particular, we propose factoring an undirected graph $\bm{A} \in \mathbb{R}_+^{n \times n}$ into two components: a bipartite graph $\bm{V} \in \mathbb{R}_+^{n \times m}$, which connects the $n$ original nodes to $m$ latent nodes, where $m < n$, and a smaller undirected graph $\bm{W} \in \mathbb{R}_+^{m \times m}$, which is a graph on the latent nodes. Intuitively, this factorization approximates taking one step of a random walk on $\bm{A}$ as a three step procedure: first taking one random step on $\bm{V}$ from the original nodes to the latent nodes, then one random step within the latent graph $\bm{W}$, and finally one random step on $\bm{V}$ back from the latent to the original nodes:
\begin{equation}\label{eqn:threestepintro}
    \pi(\bm{A}) \approx \pi(\bm{V}) \, \pi(\bm{W}) \, \pi(\bm{V}^\top) ,
\end{equation}
where $\pi$ denotes dividing each row of a matrix by its sum, yielding the random walk transition matrix corresponding to the adjacency matrix. Figure~\ref{fig:diagram} illustrates this process. As we discuss in Section~\ref{sec:method}, this model permits a differentiable parametrization, allowing for fitting via gradient descent on a simple cross-entropy loss. Further, we can ensure that the transition matrix on the right-hand side is reversible, meaning that it corresponds exactly to one step of a random walk on some undirected graph $\bm{B} \in \mathbb{R}_+^{n \times n}$. Our model allows for retrieval of this $\bm{B}$ as a rank-$m$ approximation of $\bm{A}$, connecting this clustering method to graph simplification.

\begin{figure}[h]
\centering
\vspace{-0.3em}
\includegraphics[width=0.95\columnwidth]{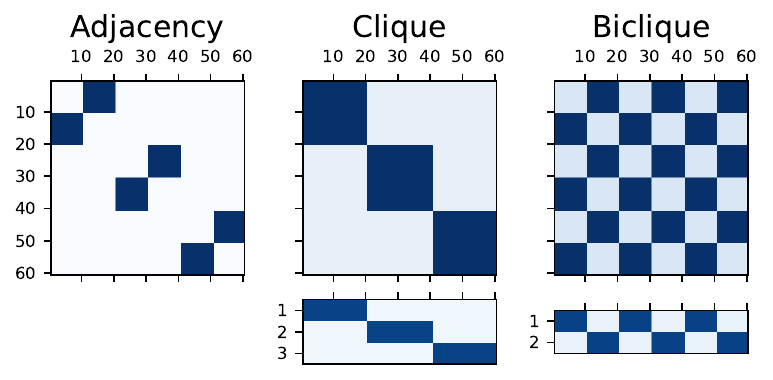}
\vspace{-0.5em}
\caption{The synthetic graph $\bm{A}$ (left) and two different clusterings (right), for which we show $\bm{B}$ (top) and $\pi(\bm{V})^\top$ (bottom).}
\label{fig:synth_new}
\vspace{-0.75em}
\end{figure}

\paragraph{Demonstrative Toy Graph.}
We construct a synthetic network that exhibits both homophily and heterophily as a concrete demonstration of how our model can adapt to both. Consider a union of 3 bicliques, each with 10 nodes in either set, for a total of 60 nodes. Figure~\ref{fig:synth_new} (left) is a plot of this graph. We can naturally cluster the nodes in at least two ways: either we find 3 homophilous clusters with most edges \emph{within} clusters (as in the min-cut task), or we find 2 heterophilous clusters with most edges \emph{across} clusters (like the max-cut task). As we discuss in Section~\ref{sec:realworld}, in our model, this corresponds to fixing one of the following latent graphs $\bm{W}$, then finding a bipartite graph $\bm{V}$ that minimizes the approximation error in Equation~\ref{eqn:threestepintro}:
\[ \bm{W}_\text{clique} = \frac{1}{3} \begin{pmatrix}
            1 & 0 & 0\\
            0 & 1 & 0\\
            0 & 0 & 1
            \end{pmatrix} \qquad
    \bm{W}_\text{biclique} = \frac{1}{2} \begin{pmatrix}
            0 & 1\\
            1 & 0
            \end{pmatrix}.\]
In Figure~\ref{fig:synth_new} (right), we show the $\pi(\bm{V})$ matrices, which map each node to a distribution over clusters, that result from both ways of clustering the graph. The one with $\bm{W}_\text{clique}$ indeed assigns each node to three clusters, corresponding to the three bicliques, so that edges occur only within clusters; whereas the one with $\bm{W}_\text{biclique}$ assigns each node to one of two clusters so as to split each biclique, such that edges occur only across clusters. 
Each method also results in a different simplified graph $\bm{B}$: the former erases the biclique structure and results in three cliques; whereas the latter erases the distinction between the three disjoint bicliques, resulting in a single large biclique. 
The latter clustering may be more useful for mining data structure in some applications, but most algorithms only allow for the former.

Our key contributions can be summarized as follows:
\vspace{-1em}
\begin{itemize}
    \item We present our \textit{latent random step} model. To our knowledge, this is the first probabilistic model for undirected graph simplification that accommodates arbitrary homophilous and heterophilous structure. 
    \item Unlike similar work, our model admits an unconstrained parametrization. Simple gradient descent (and its variants) on a natural probabilistic loss can be used to fit our model, allowing for flexible node clustering and low-rank approximation of a weighted graph.
    Modifying the latent graph $\bm{W}$ ranges the type of node clustering task through relaxations of potentially hard problems like $k$-way max-cut.
    \item We apply our model and algorithm to real-world data by simplifying weighted graphs constructed from raw orthographic and phonological data. We find that graphs with heterophilous structure naturally arise when considering the sequences of letters and phonemes in words. From these graphs, our unsupervised algorithm finds heterophilous clusters that closely align with ground-truth labels.
\end{itemize}

\section{Related Work}

Our model can be represented as an approximate graph factorization of the form $\bm{A} \approx \bm{U} \bm{W} \bm{U}^\top$, which has also been employed in some prior works. \citet{yu2005soft} present such a model for use in soft clustering and also discuss its interpretation in terms of random walks, but they focus on the case where $\bm{W}$ is diagonal, that is, the homophilous case.
Perhaps the closest model to ours is that of \citet{nourbakhsh2014matrix}, who also allow their equivalent of $\bm{W}$ to be non-diagonal.
However, their experiments do not explore non-homophilous clustering, and they do not work within a fully probabilistic framework as we do; among other differences, the loss they optimize is based on Frobenius norm rather than cross-entropy.
While these are the two models that are closest to ours, all three models fit under the umbrella of non-negative matrix factorization (NMF) for node clustering and graph simplification, for which there is other prior work~\citep{ding2008nonnegative, kuang2012symmetric}.

Our model, along with some of the other discussed models, can be seen as a very generalized variant of the well-known stochastic block model (SBM) \citep{holland1983stochastic}. The key differences are that: 1) nodes in the SBM are assigned to exactly one community, as opposed to our model's distributional assignments; and 2) the central probability matrix in the SBM gives probabilities of nodes in two communities being connected, which is slightly different from our model, wherein the central matrix gives the proportion of edges which occur between two communities; and 3) SBMs, while capable of representing heterphilous structure, are also typically studied in the context of homophilous structure. 
As suggested by use of the term `latent' states, our model can also be seen as an instance of the Hidden Markov Model~\citep{baum1966statistical} to the process of taking a random walk on a graph; unlike in most applications of HMMs, here the analyzed process is explicitly first-order, by construction.
Finally, our fitting algorithm joins much prior work as a relaxation of a computationally-hard node partitioning problem; perhaps best-known is the work of \citet{goemans1995improved}, which gives a spectral relaxation of the max-cut problem. Unlike that work, we provide no theoretical guarantees of performance, though we observe good performance in experiments. On the other hand, our framework can go well beyond max-cut to unify min-cut, $k$-way max-cut, and more, depending on how the latent graph $\bm{W}$ is set.


\section{Methodology}\label{sec:method}

As stated in Section~\ref{sec:intro}, we propose to approximately factorize an undirected graph $\bm{A} \in \mathbb{R}_+^{n \times n}$ into a bipartite graph $\bm{V} \in \mathbb{R}_+^{n \times m}$ and a smaller undirected graph $\bm{W} \in \mathbb{R}_+^{m \times m}$. We seek
\begin{equation} \label{eqn:threestep}
    \pi(\bm{A}) \approx \pi(\bm{V}) \, \pi(\bm{W}) \, \pi(\bm{V}^\top) = \pi(\bm{B}),
\end{equation}
where again $\pi$ denotes dividing each row of a matrix by its sum, yielding a random walk transition matrix. $\bm{B}$, which is a symmetric matrix in $\mathbb{R}_+^{n \times n}$, is a rank-$m$ reconstruction of $\bm{A}$ (that is, a simplified version of $\bm{A}$); like $\bm{A}$, $\bm{B}$ can be seen as an undirected, weighted graph on the original $n$ nodes.

\paragraph{Reversibility Criterion.}
%
%
We first establish a condition on $\bm{V}$ and $\bm{W}$ for the transition matrix $\pi(\bm{B})$ from Equation~\ref{eqn:threestep} to be reversible, that is, to correspond to a random step in an \emph{undirected} graph $\bm{B}$. 
This condition is crucial for fitting the model to not only yield a clustering of the nodes (given by $\pi(\bm{V})$), but also a simplified graph $\bm{B}$.
Reversibility is satisfied iff there exists a diagonal matrix $\bm{D_B} \in \mathbb{R}_+^{n \times n}$ for which the product $\bm{D_B} \, \pi(\bm{B})$ is a symmetric matrix $\bm{B}$. We assume that $\bm{V}$ and $\bm{W}$ are not just non-negative, but strictly positive; we will only parametrize such $\bm{V}$ and $\bm{W}$ anyway.

Let $\bm{D_V}$ and $\bm{D_{V}}'$ denote the diagonal matrices whose diagonal elements are the row-sums and column-sums of $\bm{V}$, respectively, and let $\bm{D_W}$ denote the row-sums of $\bm{W}$ (which are equivalent to the columns-sums since $\bm{W}$ is symmetric). We have reversibility if, for some $\bm{D_B}$, the following matrix is symmetric:
%
%
\begin{align*}
    \bm{B} &= \bm{D_B} \, \pi(\bm{V}) \, \pi(\bm{W}) \, \pi(\bm{V}^\top) \\
    &= \bm{D_B} \left( \bm{D_V}^{-1} \bm{V} \right) \left( \bm{D_W}^{-1} \bm{W} \right) \left( \bm{D_V}'^{-1} \bm{V}^\top \right) \nonumber \\
    &= \bm{D_B} \bm{D_V}^{-1} \left( \bm{V} \bm{D_W}^{-1} \bm{W} \bm{D_V}'^{-1} \bm{V}^\top \right).
\end{align*}
The transpose of this matrix is
\begin{align*}
    \bm{B}^\top = \left( \bm{V} \bm{D_V}'^{-1} \bm{W} \bm{D_W}^{-1} \bm{V}^\top \right) \bm{D_V}^{-1} \bm{D_B}.
\end{align*}
Note that if $\bm{D_V}' = \bm{D_W}$, then the parenthesized parts of the final expressions are equivalent. Further, with $\bm{D_B} = \bm{D_V}$, the matrix is fully equal to its transpose and is therefore symmetric. Explicitly, the matrix simplifies to the form
\begin{align*}
    \bm{B} = \bm{V} \bm{D_W}^{-1} \bm{W} \bm{D_W}^{-1} \bm{V}^\top . \label{eqn:B}
\end{align*}
Hence reversibility is satisfied if the column-sums of the bipartite graph $\bm{V}$ are equal to the degrees of the latent graph $\bm{W}$. If this condition is satisfied, the degrees $\bm{D_B}$ of the reconstructed graph $\bm{B}$, which corresponds to the transition matrix $\pi(\bm{B})$, are exactly the row-sums of $\bm{V}$.


\paragraph{Parametrization.}
We can parametrize our model using two matrices of free parameters, $\bm{W_p} \in \mathbb{R}^{m \times m}$ and $\bm{V_p} \in \mathbb{R}^{n \times m}$, to represent the latent graph $\bm{W}$ and the bipartite graph $\bm{V}$, respectively.
Let $\sigma_{\text{mat}}$ and $\sigma_{\text{col}}$ denote functions which take the softmax of a matrix over all elements and over each column. We first construct $\bm{W}$ as follows:
\begin{align*}
    \bm{W} = \sigma_{\text{mat}}\left( \bm{W_p} + \bm{W_p}^\top \right),
\end{align*}
which ensures both the positivity and the symmetry of $W$; the softmax also ensures that all entries of $\bm{W}$ sum to $1$. Let $\bm{D_W}$ be the diagonal matrix containing the degrees of $\bm{W}$. We now construct $\bm{V}$ with:
\begin{align*}
    \bm{V} = \left( \sigma_{\text{col}} (\bm{V_p}) \right) \bm{D_W},
\end{align*}
which ensures the positivity of $\bm{V}$ and the reversibility criterion, that the column-sums of $\bm{V}$ are equal to the degrees $\bm{W}$.
Note that while we provide the full parametrization for generality, in the experiments in this paper, we fix $\bm{W}$ and find $\bm{V}$, so $\bm{W_p}$ is not used.

\paragraph{Fitting.}

We can fit this model via gradient descent on a simple and natural cross-entropy loss:
\begin{equation} \label{eqn:loss}
    L = -\sum\nolimits_{i,j \in [n]} \left( \bar{\bm{A}}_{ij} \log\left( \bar{\bm{B}}_{ij} \right) \right),
\end{equation}
where an overline denotes dividing a matrix by the sum of all of its elements. This loss views the adjacency matrix of the original graph $\bm{A}$ and that of the reconstructed graph $\bm{B}$ as probability distributions over pairs of nodes. Minimizing it tries to place more mass in $\bm{B}$ among node pairs which correspond to edges in $\bm{A}$. 
We additionally use an $L_2$ regularization penalty on the parameters.


Our implementation uses PyTorch~\cite{NEURIPS2019_9015} for automatic differentiation and minimizes the loss using the SciPy~\cite{scipy} implementation of the L-BFGS~\cite{liu1989limited,zhu1997algorithm} algorithm with default hyperparameters. The free parameters are initialized uniformly at random on $(-10^{-2},+10^{-2})$. The regularization term for the loss is set to $10^{-1}$ times the mean squared norm of the free parameters.


\vspace{-0.75em}
\section{Experiments on Real-World Networks}\label{sec:realworld}

To illustrate the power of our model and algorithm, we perform some experiments on real-world datasets made from English-language orthographic (spelling) and phonological (pronunciation) data.
In particular, to find structure in this data in an unsupervised manner, we construct graphs from it and factorize them using the following two latent graphs to find (soft) bipartite and tripartite clusterings:
\begin{equation*}
\bm{W}_\text{bi} = \frac{1}{2} \begin{pmatrix}
            0 & 1\\
            1 & 0
            \end{pmatrix}
\qquad
\bm{W}_\text{tri} = \frac{1}{6} \begin{pmatrix}
            0 & 1 & 1\\
            1 & 0 & 1\\
            1 & 1 & 0
            \end{pmatrix}.
\end{equation*}
Factorizing with both latent graphs corresponds to finding clusters such that intra-cluster edges are minimized: with $\bm{W}_\text{bi}$, we find two clusters, and with $\bm{W}_\text{tri}$, we find three clusters such that roughly one-third of the total edge weight is assigned to each of the three pairs of clusters. These clustering tasks can be seen as soft relaxations of the standard and \mbox{$3$-way} max-cut problems.

\vspace{-0.5em}
\paragraph{Orthographic Adjacency.}\label{sec:ortho_adj}

We construct a graph based on spellings of common English language words. The nodes of the graph correspond to the 26 letters, and the edge weights correspond to the number of times, across all common words, that two letters are directly adjacent. For the list of words, we use the 20K most frequently-used English words as determined by the Google Books Ngram Viewer~\hspace{-1pt}\footnote{Specifically, we use the list on the \href{https://github.com/first20hours/google-10000-english}{\textcolor{cyan}{google-10000-english repo}}.}.

\begin{figure}
    \centering
    \includegraphics[width=1.0\columnwidth]{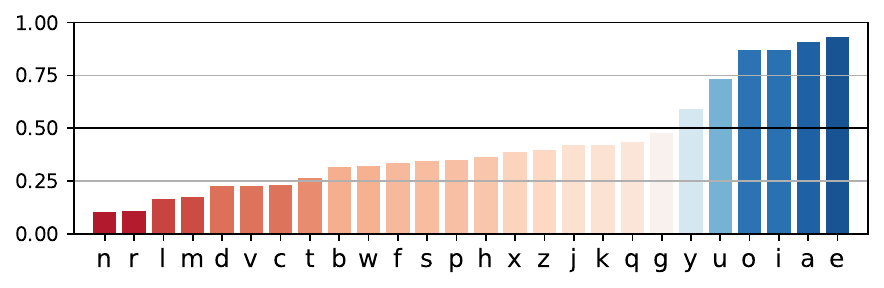}
    \vspace{-2em}
    \caption{
    Results of bipartite clustering of the orthographic adjacency graph. 
    For each letter, we plot the probability of assignment to the first of the two clusters, that is, the first column of $\pi(\bm{V})$. Letters are sorted in ascending order of this probability. 
    }
    \label{fig:letter_bipartite}
    \vspace{-0.9em}
\end{figure}

We perform a bipartite clustering of this graph using $\bm{W}_\text{bi}$ based on the intuition that the spelling of words very roughly tends to alternate between vowels and consonants.
See Figure~\ref{fig:letter_bipartite} for the results. Indeed, the resulting clustering reflects the intuition: if we convert the soft clustering $\pi(\bm{V})$ into a hard clustering by assigning each letter to the cluster for which it has higher probability, this clustering cleanly divides the letters into vowels and consonants. 
(The letter `y', which can act as both, is placed into the vowel cluster, but with the least probability among the vowels.)

\vspace{-0.5em}

\paragraph{Phonological Adjacency.}
We now construct a graph based on pronunciations of English language words. Using the same list of common words as for the orthographic data,
we convert these words to sequences of phonemes as determined by the CMU Pronouncing Dictionary~\hspace{-1pt}\footnote{This resource is hosted online at the \href{http://www.speech.cs.cmu.edu/cgi-bin/cmudict}{\textcolor{cyan}{speech.cs.cmu website}}.}. We use the NLTK API~\citep{bird2009natural} to access the dictionary. The graph is similar to the orthographic one: the nodes of this graph correspond to 39 English language phonemes, and the edge weights correspond to the number of times, across all common words in the dictionary, that two phonemes are directly adjacent.



\begin{figure}
    \centering
    \includegraphics[width=1.0\columnwidth]{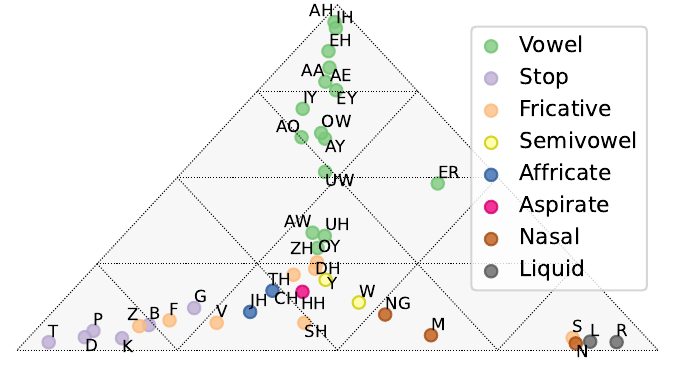}
    \vspace{-2em}
    \caption{Results of tripartite clustering of the phonological adjacency graph. This ternary plot projects the 3D categorical distributions given by the cluster affinities $\pi(\bm{V})$ onto a 2D space. Each corner corresponds to a different cluster.
    }
    \label{fig:distplot_3offdiag}
    \vspace{-2em}
\end{figure}

While applying a bipartite node clustering to phonological data also separates vowel sounds as with the orthographic data, we find that significantly more interesting structure can be extracted with a tripartite clustering, using the latent graph $\bm{W}_\text{tri}$.
See Figure~\ref{fig:distplot_3offdiag} for a plot of the resulting cluster affinities $\pi(\bm{V})$.
The clustering strongly reflects ground-truth categorizations of the phonemes. 
Most striking is that one of the clusters is dominated by vowel sounds, and that vowel, stop, and nasal/liquid sounds have high affinities for three distinct clusters. 

\vspace{-0.5em}
\section{Conclusion}
We propose our latent random step model and perform node clustering on a synthetic graph and real-world orthographic and phonological graphs, finding structure in the graphs that goes beyond typical homophilous clusterings.
The simplicity and flexibility of the model suggests several directions for extension of this work. We focus here on the setting where the latent graph $\bm{W}$ is fixed and the bipartite graph $\bm{V}$ is fit. 
%
We could instead attempt to fit both at once, yielding the full graph simplification setting. 
%
Besides this, there may be a straightforward extension to data and graphs of greater scale than we consider here: the cross-entropy loss (Equation~\ref{eqn:loss}) could be approximated by sampling of node pairs based on the weights of $\bar{\bm{A}}$, allowing for an SGD fitting algorithm.
More broadly, we hope that considering latent graphs beyond homophilous clusters can expand the applicability of node clustering, out to new problems and fields.

\bibliography{example_paper}
\bibliographystyle{icml2023}



\end{document}